\documentclass[journal]{IEEEtran}

\usepackage{graphicx}
\usepackage{times}
\usepackage{epsfig}
\usepackage{amsmath}
\usepackage{amssymb}
\usepackage{subfigure}
\usepackage{tabularx}
\usepackage{color}
\usepackage{rotating}
\usepackage{diagbox}
\usepackage{multirow}
\usepackage{bm}

\usepackage[colorlinks,linkcolor=blue]{hyperref}

\makeatletter
\newcommand*\bigcdot{\mathpalette\bigcdot@{.5}}
\newcommand*\bigcdot@[2]{\mathbin{\vcenter{\hbox{\scalebox{#2}{$\m@th#1\bullet$}}}}}
\makeatother

\ifCLASSINFOpdf
\else
\fi

\begin{document}
\title{GMLight: Lighting Estimation via Geometric Distribution Approximation}

\author{Fangneng~Zhan,
        Yingchen~Yu,
        Changgong~Zhang,
        Rongliang~Wu,
        Wenbo~Hu, \\
        Shijian~Lu$^*$,
        Feiying~Ma,
        Xuansong~Xie,
        Ling~Shao,~\IEEEmembership{Fellow,~IEEE}
\thanks{F. Zhan, Y. Yu, R. Wu, and S. Lu are with the School of Computer Science and Engineering, Nanyang Technological University, Singapore.
C. Zhang, F. Ma, and X. Xie are with the Alibaba DAMO Academy, China.
W. Hu is with the Chinese University of Hong Kong.
L. Shao is with the Inception Institute of Artificial Intelligence, UAE. 
}
\thanks{* indicates the corresponding author. Email: shijian.lu@ntu.edu.sg}
}

\markboth{IEEE Transactions on Image Processing}
{Shell \MakeLowercase{\textit{et al.}}: Bare Demo of IEEEtran.cls for IEEE Journals}
\maketitle

\begin{abstract}
Inferring the scene illumination from a single image is an essential yet challenging task in computer vision and computer graphics. Existing works estimate lighting by regressing representative illumination parameters or generating illumination maps directly. However, these methods often suffer from poor accuracy and generalization. This paper presents Geometric Mover's Light (GMLight), a lighting estimation framework that employs a regression network and a generative projector for effective illumination estimation. We parameterize illumination scenes in terms of the geometric light distribution, light intensity, ambient term, and auxiliary depth, which can be estimated by a regression network. Inspired by the earth mover's distance, we design a novel geometric mover's loss to guide the accurate regression of light distribution parameters. With the estimated light parameters, the generative projector synthesizes panoramic illumination maps with realistic appearance and high-frequency details. Extensive experiments show that GMLight achieves accurate illumination estimation and superior fidelity in relighting for 3D object insertion. The codes are available at \href{https://github.com/fnzhan/Illumination-Estimation}{https://github.com/fnzhan/Illumination-Estimation}.
\end{abstract}


\IEEEpeerreviewmaketitle

\section{Introduction}
\IEEEPARstart{E}{stimating} scene illumination from a single image has attracted increasing attention thanks to its wide spectrum of applications in image composition \cite{zhan2020aicnet}, object relighting \cite{gardner2017} in mixed reality, etc.  
As the formation of image is influenced by several factors including reflectance, scene geometry and illumination,
recovering illumination from a single image is a typical under-constrained problem.
The problem becomes more challenging as high-dynamic-range (HDR) and full-view illumination map is expected to be inferred from low-dynamic-range (LDR) image with limited field of view.

\begin{figure}[t]
\centering
\includegraphics[width=1.0\linewidth]{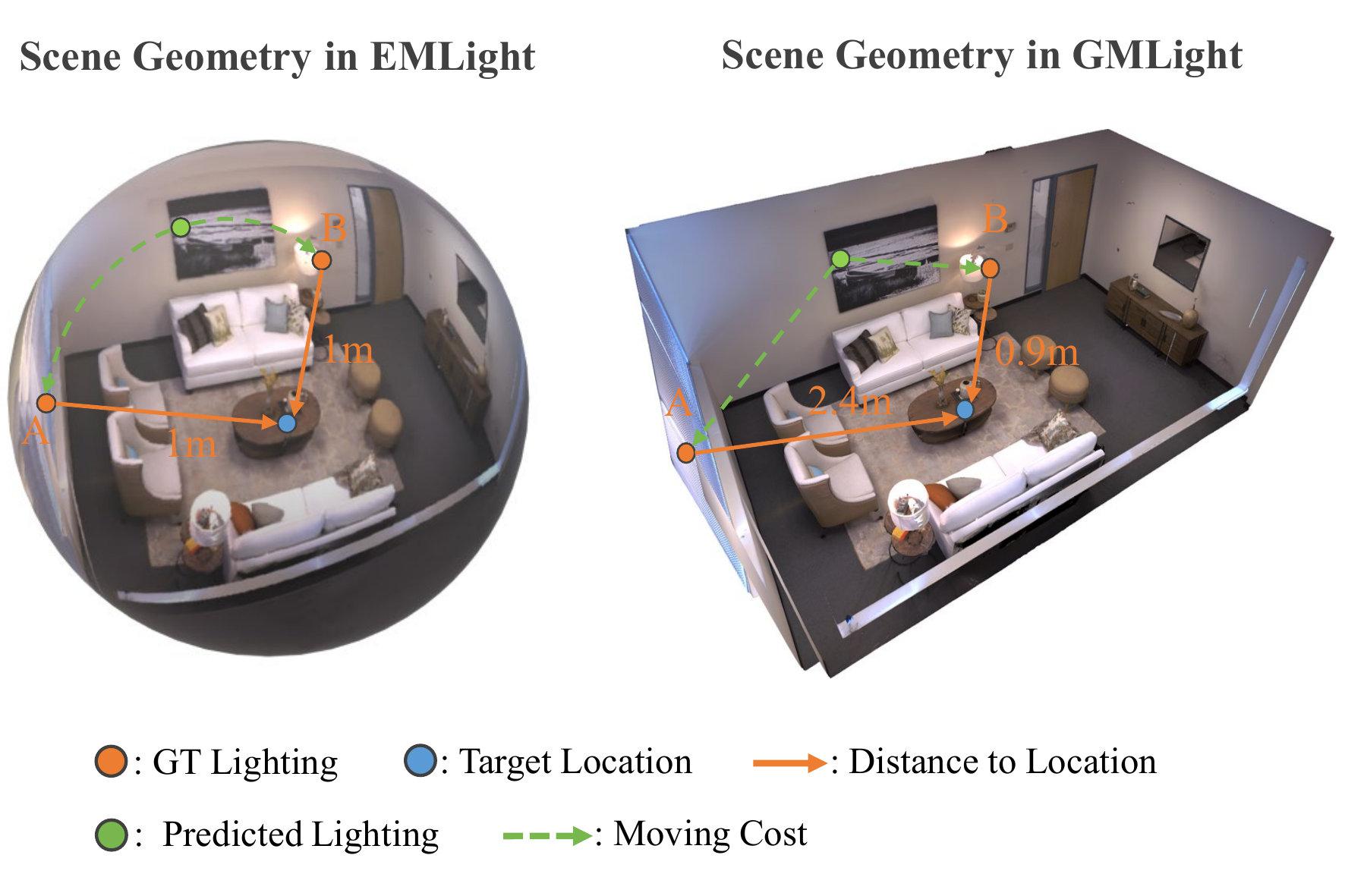}
\caption{Illustration of scene geometry in EMLight \cite{zhan2021emlight} and the proposed GMLight. GMLight regresses the illumination distribution in real geometry spaces as defined by scene depth, while EMLight simplifies the scene geometry to be a spherical surface. 
Inspired by earth mover's distance, we design a geometric mover's loss to regress the illumination distribution by minimizing the total cost required to move the predicted lighting to ground truth (GT) lighting. 
}
\label{im_intro}
\end{figure}

Early works tackle this problem by utilizing auxiliary information such as scene geometry \cite{zhang2016} and user annotation \cite{karsch2011}.
With the advancement of deep learning,
recent studies \cite{garon2019fast,gardner2019deeppara,gardner2017} attempted to estimate illumination through direct generating illumination maps or regressing the representation parameters without utilizing auxiliary information.
In particular, 
\cite{gardner2017,song2019,srinivasan2019lighthouse} aim to generate or render the illumination maps directly through neural networks.
On the other hand,
\cite{cheng2018shlight,garon2019fast} propose to recover the scene illumination by regressing the Spherical Harmonic (SH) parameters, while Spherical Gaussian (SG) representation is adopted for regression in \cite{gardner2019deeppara,li2020rendering}.
However, the regression-based methods \cite{garon2019fast,li2019spherical} often lack realistic lighting details, while the generation-based methods \cite{gardner2017,chen2019neural} may incur inaccurate prediction of light properties (e.g., light directions) and suffer from poor generalization ability.

In our previous work \cite{zhan2021emlight}, we designed EMLight which combines regression-based method and generation-based method for the prediction of environmental lighting.
Specially, the scene illumination is decomposed into a discrete distribution defined on anchor points of a spherical surface (namely spherical distribution).
Then a spherical mover's loss (SML) is designed based on earth mover's distance (EMD) \cite{emd} to regress the spherical distribution by treating the distance between anchor points as the moving cost of EMD.
However, EMLight employs a simplified spherical surface to represent the illumination scene and compute the distance between anchor points without considering scene depths.
As the formation of images are jointly determined by several intrinsic factors including the scene illumination, scene geometry, etc., over-simplification of the scene geometry will cause the lighting estimation from a single image to be unreliable. 
Besides, a simplified scene geometry also incapacitates the network in handling \textit{spatially-varying illuminations} with the method described in \cite{gardner2019deeppara}.
In this work, we propose Geometric Mover's Light (GMLight) to model light distributions in a geometric space (namely geometric distributions) as indicated by scene depth, which is closer to real-world light distributions compared with the simplified spherical distribution in EMLight as shown in Fig. \ref{im_intro}.

The proposed GMLight consists of a regression network for light parameters prediction and a generative projector for illumination map synthesis. For light parameters estimation, we propose a geometric distribution representation method to decompose illumination scene into four components: \textit{depth value}, \textit{light distribution}, \textit{light intensity}, \textit{ambient term}. Note that the last two are scalars and can be directly regressed with a naive L2 loss. Light distributions and scene depths, in contrast, are spatially-localized in scenes, and thus are not suitable to be regressed by a naive L2 loss which does not capture geometry information or SML in EMLight \cite{zhan2021emlight} which simplifies the scene geometry.
In this work, we design a Geometric Mover's Loss (GML), that regresses light distributions and scene depths with an `earth mover' in a geometric space as indicated by the scene depth. GML aims to search for an optimal plan to move one distribution to another with the minimal total cost which is defined by the moving distance in the scene.
With known scene depth, the depth value and light distribution can be effectively regressed by GML with consideration of the scene geometry.

With the estimated illumination scene parameters, the generative projector generates illumination maps with realistic appearance and details in an adversarial manner.
In detail, Spherical Gaussian function \cite{gardner2019deeppara} is adopted to map the estimated light parameters into a Gaussian map (panoramic image), which serves as the guidance in illumination map generation.
The Gaussian map can be reconstructed at each position in a scene with the knowledge of scene depth, thus achieving the estimation of \textit{spatially-varying} illuminations.
Since illumination maps are panoramic images that usually suffer from spherical distortions at different latitudes, we adopt spherical convolutions \cite{spherenet} to accurately generate panoramic illumination maps.
To provide progressive guidance in the illumination generation process, an adaptive radius strategy is designed to generate progressive Gaussian maps in a coarse-to-fine manner.
More details will be provided in Section \ref{section_generation}.

Compared with our previous work EMLight \cite{zhan2021emlight}, the contributions of this work can be summarized in three aspects. 
First, we propose a geometric distribution representation to parameterize the illumination scene for lighting estimation and enable effective estimation of \textit{spatially-varying} illumination.
Second, we introduce a novel geometric mover's loss that leverages the real scene geometry through depth values in the regression of light distribution.
Third, we design a generative projector with progressive guidance in illumination generation process. 

The rest of this paper is organized as follows. Section \ref{related} presents related works. The proposed method is then described in detail in Section \ref{method}. Experimental results are further presented and discussed in Section \ref{experiments}. Finally, concluding remarks are drawn in Section \ref{conclusion}.

\begin{figure*}[t]
\centering
\includegraphics[width=1.0\linewidth]{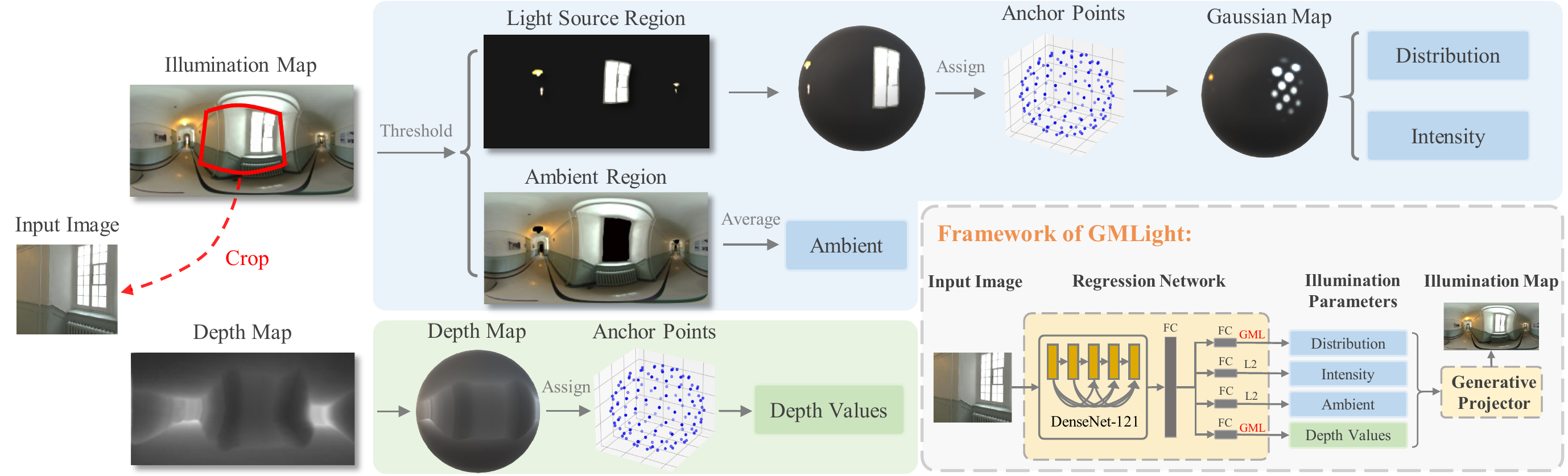}
\caption{
Illustration of the geometric distribution representation to acquire training data and the overall framework of GMLight.
Given an \textit{Illumination Map}, we first determine a \textit{Light Source Region} and \textit{Ambient Region} via thresholding.
The pixels in ambient region are averaged to obtain the ambient term and pixels in light source region are assigned to $N$ anchor points to obtain a light intensity and a light distribution as visualized by the \textit{Gaussian Map}.
Scene depths (dataset provided) are also assigned to the $N$ anchor points as ground truth.
The illumination map (scene) is thus decomposed to representative parameters including light distribution, light intensity, ambient term, and depth.
With the image cropped from the illumination map as input, the regression network in GMLight aims to regress the representative parameters, followed by a generative projector to generate the final illumination map.
FC denotes fully-connected layer, GML denotes the proposed geometric mover's loss.
}
\label{im_stru1}
\end{figure*}

\begin{figure}[ht]
\centering
\includegraphics[width=1.0\linewidth]{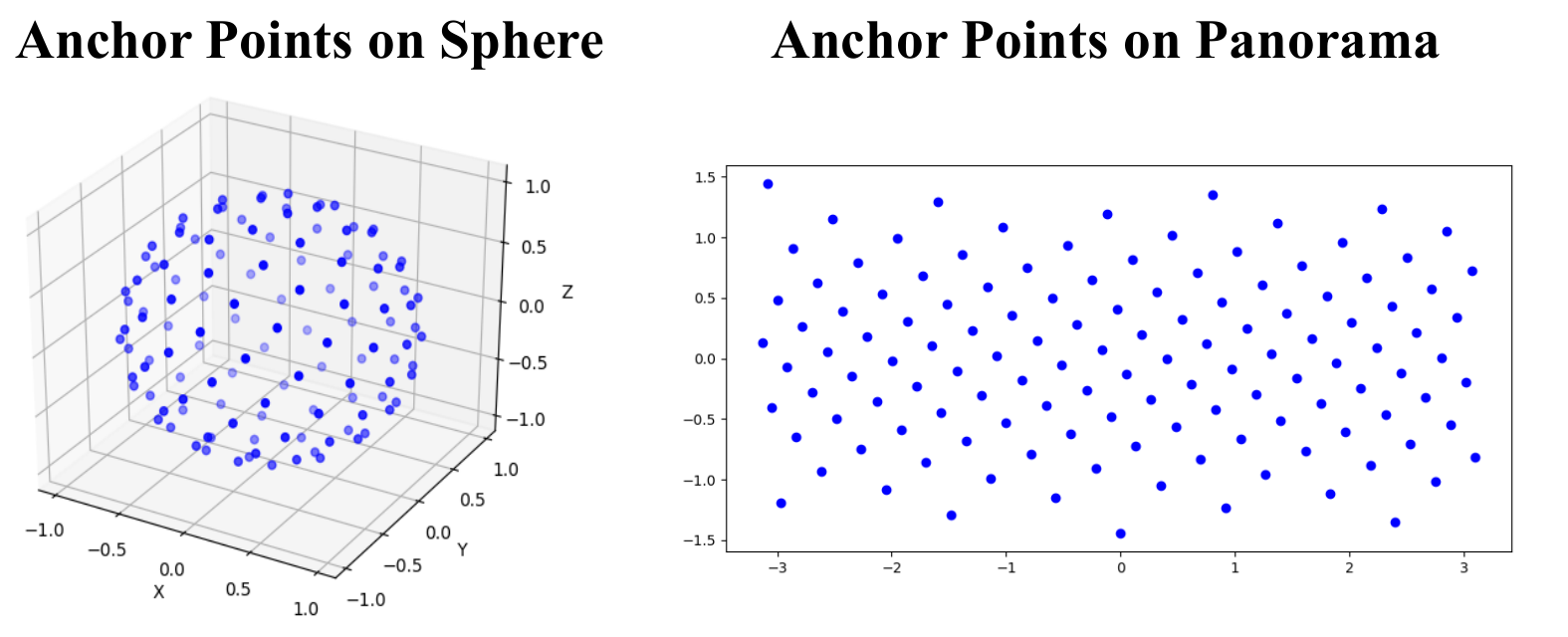}
\caption{
Visualization of 128 anchor points on a unit sphere and panorama. (pre-defined with the method described in \cite{vogel1979})
}
\label{im_anchor}
\end{figure}

\section{Related Works}
\label{related}
Lighting estimation targets to predict HDR illumination from a single image, which has been widely applied in relighting for object insertion \cite{lalonde2012,geoffroy2017,murmann2019dataset,boss2020,ngo2019reflectance,liao2019approximate,maurer2018combining} and image composition \cite{zhan2020sagan,zhan2020aicnet}.
The earlier works in this field heavily rely on auxiliary information to tackle this problem.
In particular, the scene is typically decomposed into geometry, reflectance, and illumination, and then the lighting is estimated with known scene geometry or reflectance.
For instance, 
Karsch \emph{et al.} \cite{karsch2011} acquire scene geometry through user annotations for lighting estimation. Maier et al. \cite{maier2017} employ additional depth information to estimate Spherical Harmonics (SH) representation of the illumination.  Moreover, Zhang \emph{et al.} \cite{zhang2016} utilize a multi-view 3D reconstruction of scenes to recover illumination, while Lombardi and Nishino \cite{lombardi2016} achieve illumination estimation by incorporating a low-dimensional reflectance model based on the objects with known shapes. 
Sengupta \emph{et al.} \cite{sengupta2019} jointly estimate the albedo, normals, and lighting of an indoor scene from a single image and Barron \emph{et al.} \cite{barron2015} estimate shape, lighting, and material but rely on data-driven priors to compensate for the lack of geometry information. 

Thanks to the tremendous success of deep learning, data-driven approaches have become prevalent in recent years and demonstrated the feasibility of estimating lighting without auxiliary information. They roughly fall into two main categories: 1) regression-based methods that regress the lighting representation parameters \cite{cheng2018shlight,gardner2019deeppara,li2020rendering}; 2) generation-based methods \cite{gardner2017,song2019} that directly generate illumination maps with neural networks.
For the regression-based methods,
Cheng \emph{et al.} \cite{cheng2018shlight} and Garon \emph{et al.} \cite{garon2019fast} represent the illumination with spherical harmonic (SH) parameters and predict the scene illumination by regressing the corresponding parameters.
On the other hand, Gardner \emph{et al.} Gardner \emph{et al.} \cite{gardner2019deeppara} parameterize the scene illumination into the light directions, light intensities, and light color with spherical Gaussian functions to enable the explicit regression of individual light source in the scene.
In the aspect of the generation-based method, 
Gardner \emph{et al.} \cite{gardner2017} employ a two-step training strategy to generate illumination maps. 
Song \emph{et al.} \cite{song2019} utilize a convolutional network to predict unobserved content in the environment map with predicted per-pixel 3D geometry.
Legendre \emph{et al.} \cite{legendre2019deeplight} generate the illumination maps by matching the LDR ground-truth sphere images to those rendered with the predicted illumination using image-based relighting.
Besides, 
Srinivasan \emph{et al.} \cite{srinivasan2019lighthouse} utilize volume rendering to generate incident illuminations according to a 3D volumetric RGB model of the scene.
Instead of predicting the illumination maps,
Sun \emph{et al.} \cite{sun2019} propose a framework to achieve relighting on RGB portrait images given any illumination maps.
Moreover, several works \cite{liu2020shadow,zhan2020aicnet} adopt generative adversarial network to generate shadows on RGB images directly without explicitly estimating the illumination map.

The previous lighting estimation works only adopt either the regression-based method or generation-based method alone, which tends to lose realistic lighting details or predict inaccurate light properties (e.g., light directions) respectively. 
In contrast, the proposed GMLight combines the merits of the regression-based and generation-based methods for accurate yet realistic illumination estimation.

\section{Proposed Method}
\label{method}

The proposed GMLight consists of a regression network and a generative projector that are jointly optimized.
A geometric distribution representation is proposed to parameterize illumination scenes with a set of parameters which are to be estimated by the regression network.
The predicted illumination parameters serve as guidance for the generative projector to synthesize realistic yet accurate illumination maps.

\subsection{Geometric Distribution for Illumination Representation}

As illumination maps are high-dynamic-range (HDR) images, the light intensity of different illumination maps may vary drastically, which is deleterious for the parameter regression.
We thus propose to normalize the light intensity of illumination maps and design a novel geometric distribution method for illumination representation.

As illustrated in Fig. \ref{im_stru1}, the geometric distribution illumination representation decomposes the scene illumination into four parameters including light distribution $P$, light intensity $I$, ambient term $A$, and depth $D$.
With a HDR illumination map, we first separate the light sources from the full image since the light sources in a scene play the most critical role for relighting.
Consistent with \cite{gardner2019deeppara}, the light sources are selected as top 5\% of pixels with the highest values in the HDR illumination map. The light intensity $I$ and the ambient term $A$ can then be determined by the sum of all light-source pixels and the average of remaining pixels, respectively. 
To formulate discrete distributions, we generate $N$ anchor points (using the method described in \cite{vogel1979}) that are uniformly distributed on a unit sphere as illustrated in Fig. \ref{im_anchor}.
Light-source pixels are assigned to their corresponding anchor point based on the minimum radian distance, and further determine the light-source value of the anchor point by summing all its affiliated pixels.
Afterward, the value of light-source pixels is normalized by the intensity $I$ so that the $N$ anchor points on the unit sphere form a discrete distribution (i.e. the light distribution $P$). Since Laval Indoor HDR dataset \cite{gardner2017} provides pixel-level depth annotations, the depth $D$ at each anchor point can be determined by averaging depth values of all the pixels that are assigned to the anchor point (similarly based on the minimum radian distance).

With a limited-view image cropped from the illumination map to server as the network input, the illumination parameters and the original illumination map serve as the ground truth for the regression network and generative projector, respectively.

\subsection{Regression of Illumination Parameters}
\begin{figure}[t]
\centering
\includegraphics[width=1.0\linewidth]{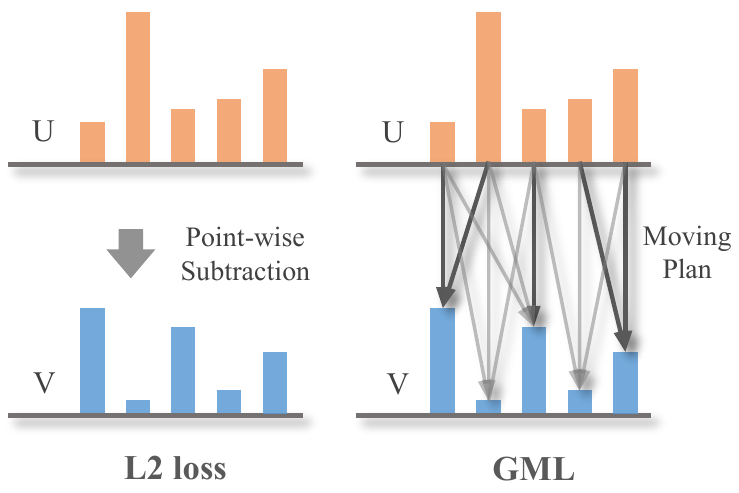}
\caption{
Comparison between L2 loss and the proposed geometric mover's loss (GML) for computing the distance between distributions $U$ and $V$. L2 loss is computed through a point-wise subtraction which cannot leverage the proximity of geometric distribution.
Instead, the geometric mover's loss aims to find the moving plan of minimal cost with consideration of the scene geometry which is correlated with the moving cost.
The thickness of connecting lines denotes the amount of probability moved between two points.
}
\label{im_gml}
\end{figure}

In the regression network, four branches with shared backbone network are adopted to regress the four sets of parameters, respectively.
The light intensity $I$ and ambient term $A$ are scalars which can be regressed with a naive L2 loss.
However, the light distribution $P$ and depth values $D$ are vectors of $N$ dimension that are spatially localized on a sphere, and the naive L2 loss cannot effectively utilize the proximity of the geometric distribution and the property of the standard distribution (for light distribution $P$, the summation of all anchor point values is equal to one).
Therefore, we propose a geometric mover's loss based on earth mover's distance \cite{emd} to effectively measure the discrepancy between distributions with consideration of the scene geometry as shown in Fig. \ref{im_gml}.

\vspace{5pt}

\begin{figure}[t]
\centering
\includegraphics[width=1.0\linewidth]{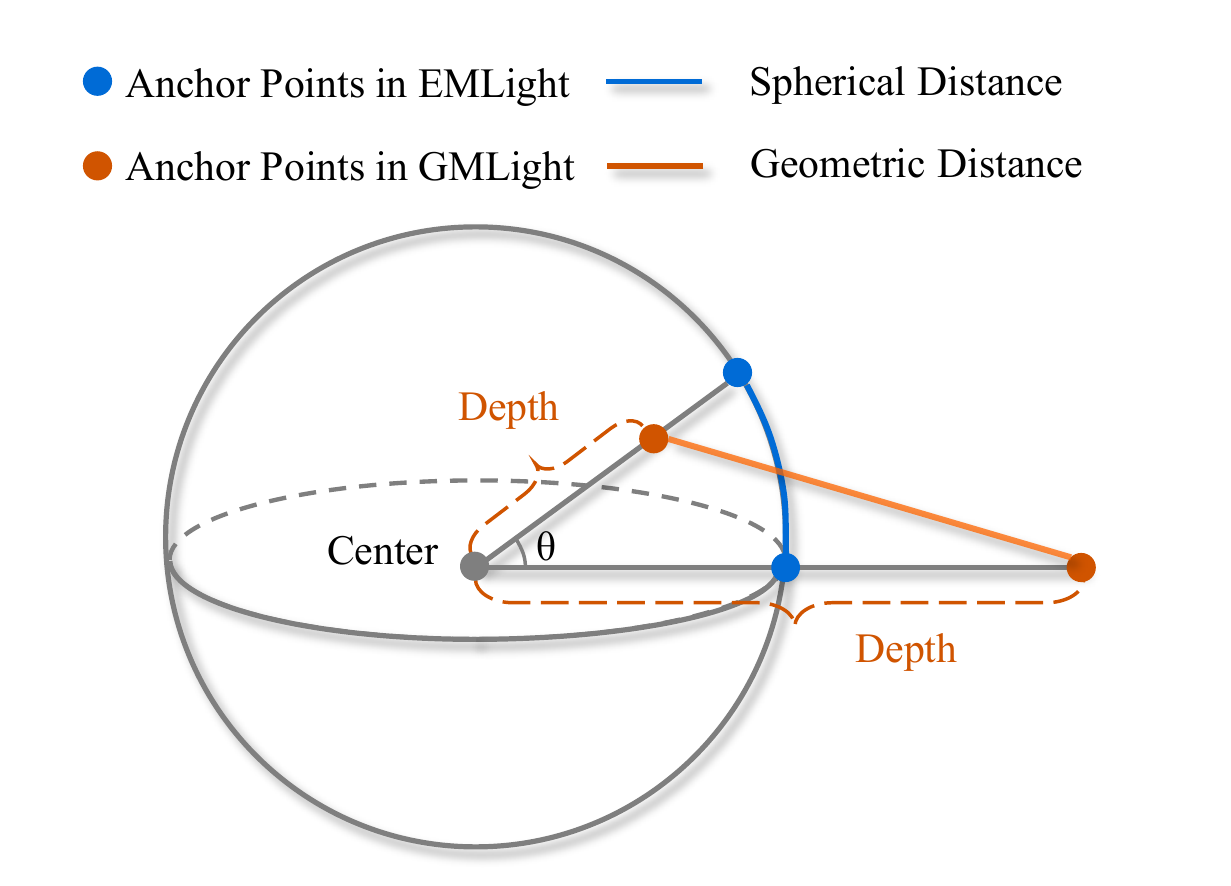}
\caption{
Geometric distances in GMLight and spherical distances in EMLight. Spherical distances in EMLight assume that anchor points are distributed on a spherical surface which neglects the complex geometry of real scenes. Geometric distances in GMLight instead capture real geometries of anchor points by using scene depths, i.e., the distance from the anchor point to the spherical center. Since the spherical angle $\theta$ between anchor points is known \cite{vogel1979}, geometric distances between anchor points can be computed.
}
\label{im_cost}
\end{figure}

\textbf{Geometric Mover's Loss:}
Our previous work EMLight proposes a spherical earth mover's loss (SML) to achieve geometry-aware regression.
However, the SML over-simplifies the scene geometry to be a spherical surface which ignores the complex geometry of real scenes.
As an image is produced as a result of the scene illumination, scene geometry, etc., ignoring the scene geometry will deteriorate the inverse process of inferring illumination from a single image. 
We thus propose a geometric mover's loss (GML) which effectively leverages the real scene geometry through depth values.
To derive the proposed GML, two discrete distributions $U$ and $V$ with $N$ points are defined as shown in Fig. \ref{im_gml}.
According to earth mover's distance, GML can be defined as the minimum total cost to convert distribution $U$ to distribution $V$, where the cost is measured by the product of the amounts of value (or `earth') to be moved and the distance to be moved.
Thus, a moving plan matrix $T$ and a cost matrix $C$ both of size $(N, N)$ are defined, where entries $T_{ij}$ and $C_{ij}$ denote the amounts of moved `earth' between point $U_{i}$ and point $V_j$ and the unit cost of moving $U_{i}$ to $V_{j}$, respectively.
In our previous spherical mover's distance, the unit cost between point $U_{i}$ and point $V_{j}$ is measured by their radian distance along the unit sphere as shown in Fig. \ref{im_cost}, which is unable to take advantage of the real scene geometry in regression.
We thus propose to facilitate the scene geometry through a geometric distance as determined by the scene depth.
As shown in Fig. \ref{im_cost}, the distance of a GMLight anchor point to the center is measured by a depth value $D$ instead of the radius, thus effectively reflecting the real geometry of the scene.
The geometric distance (or unit cost) between anchor points $U_{i}$ and $V_{j}$ can be computed according to their depth values $D_i$, $D_j$ and their spherical angle $\theta$, as follows:
\begin{equation}
    C_{ij} = {D_i}^2 + {D_j}^2 - 2 \ D_i \ D_j \ \cos(\theta) \ .
\end{equation}
With the defined transportation plan matrix $T$ and cost matrix $C$, GML can be formulated as the minimum total cost for the transport between $U$ and $V$:
\begin{equation}
\begin{split}
& GML = \mathop{\min}\limits_{T} (\sum_{i=1}^{N} \sum_{j=1}^{N} C_{ij} T_{ij}) = \mathop{\min}\limits_{T} \langle C, T \rangle \\
& subject \ to \quad T\cdot \vec{1} = U, \quad T^\top \cdot \vec{1} = V \ . \\ 
\end{split}
\label{formula_emd}
\end{equation}

\begin{figure*}[t]
\centering
\includegraphics[width=1.0\linewidth]{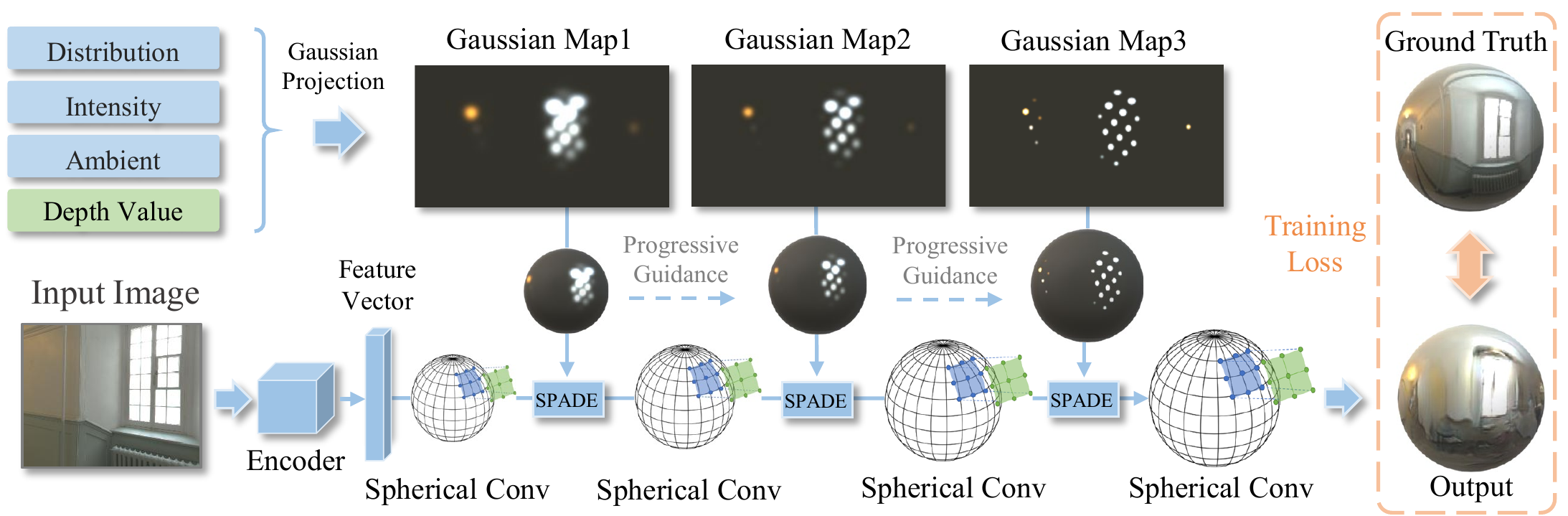}
\caption{
Structure of the generative projector. \textit{Input Image} is fed to an \textit{Encoder} to produce a \textit{Feature Vector} for the ensuing multi-stage spherical convolution.
Progressive guidance (e.g., \textit{Gaussian Map1}, \textit{Gaussian Map2}, \textit{Gaussian Map3}) are acquired from the regressed parameters (distribution, intensity, ambient, depth) via Gaussian projection with different radius parameters, and are injected into the multi-stage generation process to synthesize the \textit{Output} illumination map. \textit{SPADE} denotes spatially adaptive normalization \cite{park2019spade} for feature injection. 
}
\label{im_stru2}
\end{figure*}

On the other hand, different from the illumination distributions, the depth values of anchor points don't form a standard distribution (the sum of all depth values is not constant). Thus, we introduce an unbalanced setting of GML for the regression of depth values.
Specifically, it is handled by introducing a relaxed version of classical earth mover's distance, namely unbalanced earth mover's distance \cite{chizat2016uot}. It aims to determine an optimal transport plan between measures (depth values) of different total masses. We formulate unbalanced GML by replacing the `Hard' conservation of masses in (\ref{formula_emd}) by a `Soft' penalty with a divergence metric. An unbalanced GML as denoted by $GML^{u}$ can thus be formulated as follows:
\begin{equation}
\begin{split}
 & GML^{u} = \mathop{\min}\limits_{T} \left [ \langle C, T \rangle +  {\rm KL} (T \vec{1} | U) +  {\rm KL} (T^{\top} \vec{1} | V) \right ] \\
\end{split}
\label{formula_uemd}
\end{equation}
where ${\rm KL}$ is the Kullback-Leibler divergence which is defined as:
\begin{equation}
{\rm KL} (a||b) = \sum_{i = 1}^{n} a_{i} \log (\frac{a_{i}}{b_{i}}) - a_{i} + b_{i} \ .
\end{equation}

Grounded in well-studied entropic regularization \cite{cuturi2013sinkhorn} for differentiable optimization, the entropic version of GML (\ref{formula_emd}) can be formulated as below:
\begin{equation}
\begin{split}
& GML = \mathop{\min}\limits_{T} \langle C, T \rangle - \epsilon H(T) \\
& subject \ to \quad T\cdot \vec{1} = U, \quad T^\top \cdot \vec{1} = V \ , \\
\end{split}
\end{equation}
where $H(T)$ is the entropic term as defined by $H(T) = - \sum_{i=1}^{N} \sum_{j=1}^{N} T_{ij} \log T_{ij}$, $\epsilon$ is the regularization coefficients denoting the smoothness of the transportation plan matrix $T$. In our model, $\epsilon$ is empirically set to 0.0001.
The unbalanced GML (\ref{formula_uemd}) can be regularized similarly by adding the entropic term $H(T)$.
Then the entropic version of the problem (\ref{formula_emd}) and (\ref{formula_uemd}) can be solved in a differentiable yet efficient way through Sinkhorn iteration \cite{cuturi2013sinkhorn} in training.

The proposed geometric mover's loss makes the regression sensitive to the global geometry which effectively penalizes the spatial discrepancy between the predicted distribution and ground-truth distribution. Besides, the geometric mover's loss is smoother than L2 loss which enables more stable optimization in training process.

\subsection{Illumination Generation}
\label{section_generation}
The regressed light parameters indicate accurate light properties including the light intensity, ambient and light distributions.
To project these parameters into illumination maps, we propose a novel generative projector based on generative adversarial networks (GANs) \cite{goodfellow2014gan} as illustrated in Fig. \ref{im_stru2}.
Specifically, the light distribution $P$, light intensity $I$, and ambient term $A$ are firstly projected to Gaussian map $M$ through spherical Gaussian function \cite{gardner2019deeppara} as follows:
\begin{equation}
M = \sum_{i=1}^{N} v_i * \exp \frac{o_{i}*u-1}{s} + A \ ,
\label{gaussian}
\end{equation}
where $N$ is the number of anchor points, 
$v_{i} = P_{i}*I$ and $o_{i}$ denote the RGB value and the direction of an anchor point respectively, 
$u$ is a unit vector on the sphere and $s$ is the angular size which is set to 0.0025 empirically.
Then the illumination generation can be formulated as an image-to-image translation task conditioned on the Gaussian map.

Fig. \ref{im_stru2} illustrates the image translation process, where the input image is encoded into a feature vector to serve as the network input.
To provide effective guidance in the generation process, the Gaussian map is injected into generation process through a spatially adaptive normalization (SPADE) \cite{park2019spade} as shown in Fig. \ref{im_stru2}.
As the illumination map is a panoramic image where pixels are stretched at different latitudes, the vanilla convolution suffers from heavy distortions around the polar regions of the illumination map.
To address this, Benjamin \emph{et al.} \cite{spherenet} propose to encode the invariance against latitude distortions into convolutional neural networks by adjusting the locations of convolution operations, as known as spherical convolution.
We thus employ the spherical convolution (Spherical Conv) in the generative projector to synthesize panoramic illumination map which effectively mitigates the distortions of different latitudes on the illumination map.

In EMLight \cite{zhan2021emlight}, the same conditional Gaussian map is injected into the generation process at different levels.
Ideally, coarse Gaussian map guidance is expected in low generation layers to learn the overall illumination condition, while fine Gaussian map is expected in the high layers to indicate the accurate illumination distribution.
Thus, we design an adaptive radius strategy in the Gaussian function to generate coarse-to-fine Gaussian maps for different generation layers, as shown in `Gaussian Map1', `Gaussian Map2', and `Gaussian Map3' in Fig. \ref{im_stru2}.

\vspace{5pt} 
\textbf{Spatially-Varying Projection:}
Spatially-varying illumination prediction aims to recover the illumination of different positions in a scene from a single image. As there are no annotations for spatially-varying illumination in the Laval Indoor HDR dataset, we are unable to train a spatially varying model directly.
Previous research \cite{gardner2019deeppara} proposed to incorporate depth values into the projection to approximate the effect of spatially-varying illumination.
We follow a similar idea with \cite{gardner2019deeppara} to achieve the estimation of spatially varying illumination, as described below.

The Gaussian map is constructed through a spherical Gaussian function, as shown in Eq. (\ref{gaussian}).
When we move the insertion position by $\nabla o$, the new direction of the anchor point $i$ can be denoted by $o_{i} + \nabla o$. The depth of the original insertion position and the new position are $l_{i}$ and $l_{i} + \nabla l$, which can be obtained from the predicted depth value of $N$ anchor points.
The light intensity at the new insertion position can thus be approximated by $v_{i} * \frac{l_i}{l_i + \nabla l}$, and the Gaussian map $M_{\nabla}$ of the new insertion position can be constructed as follows:
\begin{equation}
M_{\nabla} = \sum_{i=1}^{N} v_i (\frac{l_i}{l_i + \nabla l}) * \exp \frac{(o_{i} + \nabla o)*u-1}{s} + A
\end{equation}

The Gaussian map is then fed into the generative projector to synthesize the final illumination map.
Fig. \ref{im_sample} illustrates several samples of predicted Gaussian maps, generated illumination maps, visualized intensity maps and the corresponding ground truth.
Fig. \ref{im_spatial} illustrates the generated spatially-varying illumination maps at different insertion positions.

\vspace{5pt}
\textbf{Loss Functions:}
Several losses are employed in the generative projector to yield realistic yet accurate illumination maps.
For clarity, the input Gaussian map, ground-truth illumination map, and generated illumination map are denoted by $x$, $y$, and $x'$, respectively.

To synthesize realistic illumination maps, a discriminator with Patch-GAN \cite{isola2017pixel2pixel} structure is included to impose an adversarial loss as denoted by $\mathcal{L}_{adv}$.
Different from the vanilla GAN discriminator which discriminates real from faked at image level,  Patch-GAN discriminator has a fully convolutional architecture and only conduct the discrimination at the scale of local patches.
Image discrimination results across all the patches are averaged to provide the ultimate output of discriminator. 
The Patch-GAN achieves better synthesis quality and detail generation in image translation tasks \cite{isola2017pixel2pixel} compared with vanilla GAN discriminator.

Then a feature matching loss $\mathcal{L}_{feat}$ is introduced to stabilize the training by matching the intermediate features of the discriminator between the generated illumination map and the ground truth: 
\begin{equation}
    \mathcal{L}_{feat} = \sum_{l} \lambda_{l} ||D_{l}(x, x') - D_{l}(x, y) ||_{1} \ ,
\end{equation}
where $D_{l}$ represents the activation of layer $l$ in the discriminator and $\lambda_{l}$ denotes the balanced coefficients.
Targeting to yield similar illumination distribution regardless of the absolute intensity,
a cosine similarity is employed to measure the distance between the generated illumination map and ground truth as follows:
\begin{equation}
    \mathcal{L}_{cos} = 1 - Cos(x', y),
\end{equation}
Then, the generative projector is optimized under the following objective:
\begin{equation}
    \mathcal{L} = \mathop{\min}\limits_{G} \mathop{\max}\limits_{D} ( \lambda_{1} \mathcal{L}_{feat} + \lambda_{2}   \mathcal{L}_{cos} + \lambda_{3} \mathcal{L}_{adv}) \ .
\end{equation}
where $\lambda$ balances the loss terms.
As the regression network and generative projector are both differentiable, the whole framework can be optimized end-to-end.

\section{Experiments}
\label{experiments}

\subsection{Datasets and Experimental Settings}
We benchmark GMLight with other SOTA lighting estimation models on the Laval Indoor HDR dataset \cite{gardner2017}.
To acquire paired data for model training, 
we crop eight images with limited fields of view from each panorama in the Laval Indoor HDR dataset, which finally produces 19,556 training pairs for our experiments. 
Specifically, the image warping operation as described in \cite{gardner2017} is applied to each panorama to mimic the light locality for indoor scenes.
In our experiments, 200 images are randomly selected as the testing set and the rest images are used for training.
In addition to the Laval Indoor HDR dataset, we also qualitatively evaluate GMLight on the dataset \footnote{https://lvsn.github.io/fastindoorlight/} introduced in \cite{garon2019fast}. 

Following \cite{gardner2019deeppara} and \cite{garon2019fast}, we use DenseNet-121 \cite{huang2017densely} as backbone in regression network.
The default size $N$ of the discrete light distribution is 128.
The sizes of the input image and reconstructed illumination map are $192 \time 256$ and $128 \times 256$.
The detailed network structure of the generative projector is provided in the supplementary material. We implemented GMLight in PyTorch and adopted the Adam algorithm \cite{kingma2014adam} as the optimizer that employs a learning rate decay mechanism (the initial learning rate is 0.001). The network is trained on two NVIDIA Tesla P100 GPUs with a batch size of 4 for 100 epochs.

\begin{figure}[ht]
\centering
\includegraphics[width=1.0\linewidth]{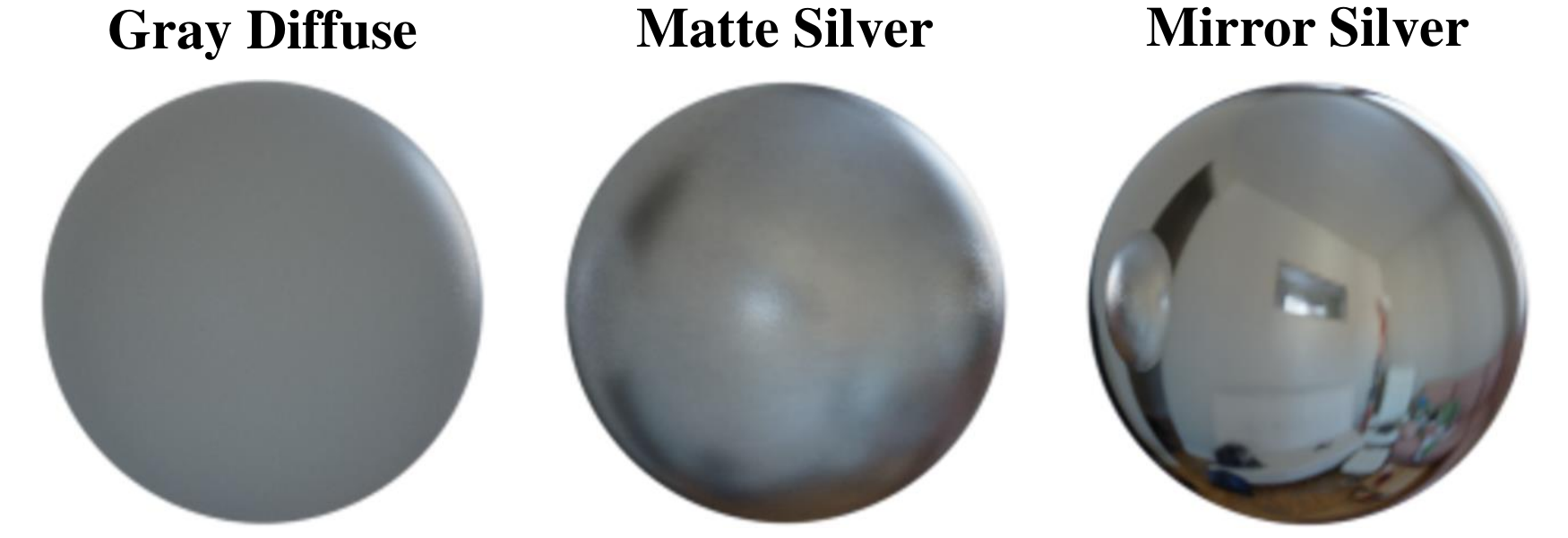}
\caption{
The scenes in evaluations are spheres of three types of materials including diffuse gray, matte silver, and mirror silver.
}
\label{im_ball}
\end{figure}

\renewcommand\arraystretch{1.2}
\begin{table*}[t]
\small 
\caption{
Comparison of GMLight with several state-of-the-art lighting estimation methods. The evaluation metrics include the widely used RMSE, si-RMSE, angular error, user study and GMD. D, S, and M denote diffuse, matte silver, and mirror materials of the rendered objects, respectively.
}
\renewcommand\tabcolsep{5pt}
\centering 
\begin{tabular}{l|ccc|ccc|ccc|ccc|p{1.2cm}<{\centering}} 
\hline
\multirow{2}{*}{\textbf{Metrics}} & 
\multicolumn{3}{c|}{\textbf{RMSE $\downarrow$}} & 
\multicolumn{3}{c|}{\textbf{si-RMSE} $\downarrow$ } & 
\multicolumn{3}{c|}{\textbf{Angular Error} $\downarrow$} &
\multicolumn{3}{c|}{\textbf{User Study} $\uparrow$} & 
{\textbf{GMD} $\downarrow$} \\
\cline{2-14}
 & D & S & M & D & S & M& D & S & M & D & S & M & N/A \\\hline

\textbf{Gardner \emph{et al.} \cite{gardner2017}} & 0.146 & 0.173 & 0.202   & 0.142 & 0.151 & 0.174   & 8.12$^{\circ}$ & 8.37$^{\circ}$ & 8.81$^{\circ}$     & 28.0\% & 23.0\% & 20.5\%    & 6.842    \\

\textbf{Gardner \emph{et al.} \cite{gardner2019deeppara}} & 0.084 & 0.112 & 0.147   & 0.073 & 0.093 & 0.119   & 6.82$^{\circ}$ & 7.15$^{\circ}$ & 7.22$^{\circ}$   & 33.5\% & 28.0\% & 24.5\%  & 5.524   \\

\textbf{Li \emph{et al.} \cite{li2019spherical}} & 0.203 & 0.218 & 0.257   & 0.193 & 0.212 & 0.243   & 9.37$^{\circ}$ & 9.51$^{\circ}$ & 9.81$^{\circ}$   & 25.0\% & 21.5\% & 17.5\%   & 7.013  \\

\textbf{Garon \emph{et al.} \cite{garon2019fast}}  & 0.181 & 0.207 & 0.249     & 0.177 & 0.196 & 0.221      & 9.12$^{\circ}$ & 9.32$^{\circ}$ & 9.49$^{\circ}$      & 27.0\% & 22.5\% & 19.0\%      & 7.137    \\

\textbf{EMLight \cite{zhan2021emlight}} & 0.062 & 0.071 & 0.089       & 0.043 & 0.054 & 0.078      & 6.43$^{\circ}$ & 6.61$^{\circ}$ & 6.95$^{\circ}$     & 40.0\% & 35.0\% & 25.0\%  & 5.131  \\

\textbf{NeedleLight \cite{zhan2021sparse}} & 0.072 & 0.074 & 0.091       & 0.051 & 0.062 & 0.084      & 6.61$^{\circ}$ & 6.78$^{\circ}$ & 7.04$^{\circ}$     & \textbf{43.0\%} & 33.5\% & 29.0\%  &  \textbf{4.213} \\

\hline

\textbf{GMLight} & \textbf{0.051} & \textbf{0.064} & \textbf{0.078}       & \textbf{0.037} & \textbf{0.049} & \textbf{0.074}      & \textbf{6.21$^{\circ}$} & \textbf{6.50$^{\circ}$} & \textbf{6.77$^{\circ}$}     & 42.0\% & \textbf{35.5\%} & \textbf{31.0\%}  & 4.892 \\
\hline
\end{tabular}
\label{tab_compare}
\end{table*}

\begin{figure*}[ht]
\centering
\includegraphics[width=1.0\linewidth]{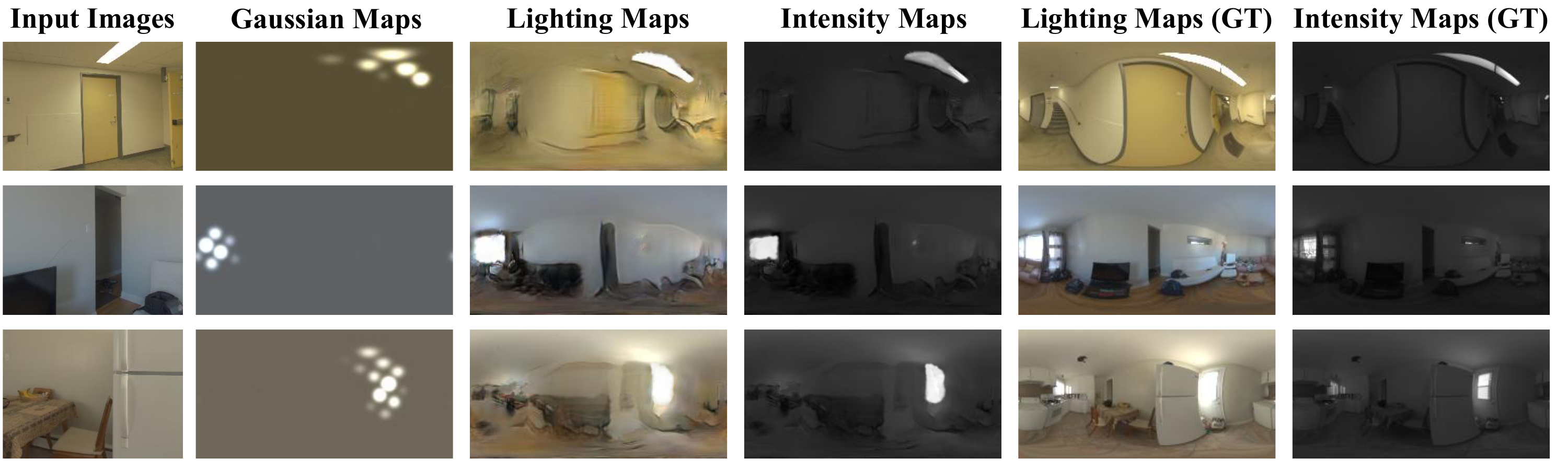}
\caption{
Illustration of GMLight illumination estimation. For the \textit{Input Images} in column 1, columns 2, 3, and 4 show the estimated \textit{Gaussian Maps} (based on the regressed illumination parameters), \textit{Lighting Maps} and \textit{Intensity Maps}, respectively. Columns 5 and 6 show the ground truth of the corresponding intensity maps and illumination maps (i.e. \textit{Lighting Maps (GT)} and \textit{Intensity Maps (GT)}), respectively.
}
\label{im_sample}
\end{figure*}

\begin{figure*}[ht]
\centering
\includegraphics[width=1.0\linewidth]{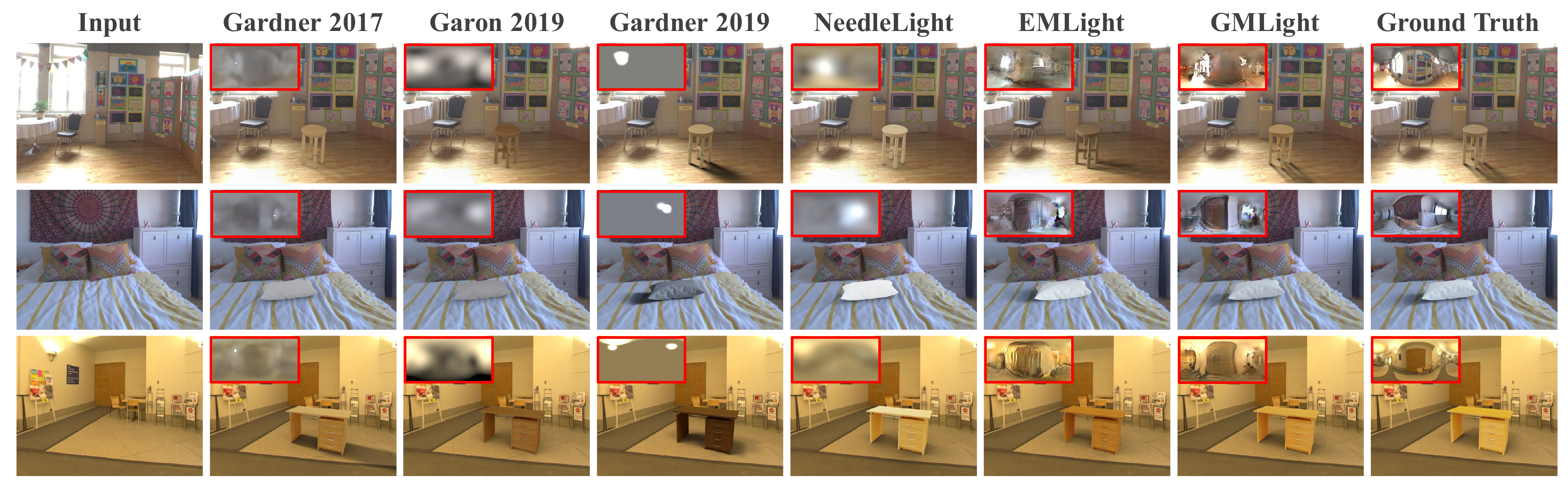}
\caption{
Qualitative comparison of GMLight with state-of-the-art methods: With illumination maps predicted by different methods (shown at the top-left corner of each rendered image), the rendered objects demonstrate different light intensities, colors, shadows, and shades.
}
\label{im_render}
\end{figure*}

\begin{figure}[ht]
\centering
\includegraphics[width=1.0\linewidth]{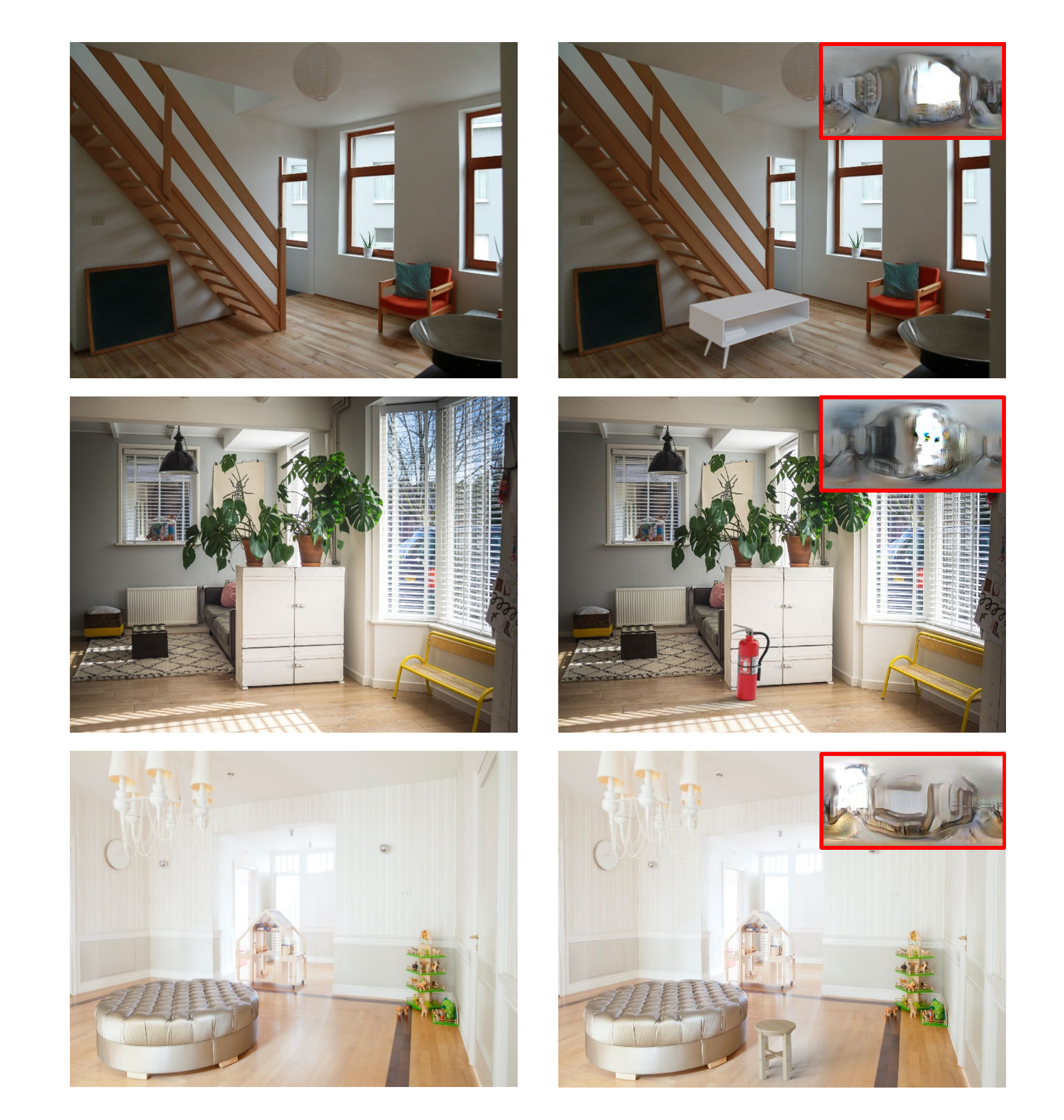}
\caption{
Object relighting over sample images from the Internet. For sample images in Column 1, our GMLight estimates illumination maps (highlighted by red boxes) automatically and applies them to relight virtual objects realistically as shown in Column 2.
}
\label{im_render2}
\end{figure}

\subsection{Evaluation Method and Metrics}

For accurate and comprehensive evaluation of the model performance,
we create a virtual scene consisting of three spheres made of gray diffuse, matte silver and mirror as illustrated in Fig. \ref{im_ball}.
By comparing the sphere images rendered (by Blender \cite{blender}) with the ground-truth illumination and the predicted illuminations, the illumination estimation performance can be effectively measured.
Several metrics are employed to compare the rendered images including root mean square error (RMSE) which mainly evaluates the accuracy of light intensity, scale-invariant RMSE (si-RMSE) and RGB angular error \cite{legendre2019deeplight} which mainly evaluate the accuracy of light directions.
Besides, we also perform crowdsourcing user study through Amazon Mechanical Turk (AMT) to assess the realism of images rendered with illumination maps predicted by different methods.
Two images rendered by the ground truth and each compared method are shown to 20 users who are asked to pick a more realistic image. The score is the percentage of rendered images that are deemed more realistic than the ground-truth rendering.
The testing scene of three spheres is mainly used for \textit{quantitative evaluation}.
We also design 25 virtual 3D scenes on the testing set for object insertion to evaluate the \textit{qualitative performance} in various scenes.

Besides, we introduce a Geometric Mover's distance (GMD) based on the geometric mover's loss as described in (\ref{formula_emd}) to measure the discrepancy between light distributions of illumination maps as below:
\begin{equation}
\begin{split}
 & GMD = \mathop{\min}\limits_{T} \left [ \langle C, T \rangle \right ], \quad T\cdot \vec{1} = U, \; T^\top \cdot \vec{1} = V \ , \\
\end{split}
\label{formula_gmd}
\end{equation}
where $U$ and $V$ are the normalized illumination maps, and the pixels in the maps form geometric distributions.
The GMD metric is sensitive to the scene geometry with the cost matrix $C$, thus achieving a more accurate evaluation of illumination distribution (or directions) compared with si-RMSE.

\begin{figure*}[t]
\centering
\includegraphics[width=1.0\linewidth]{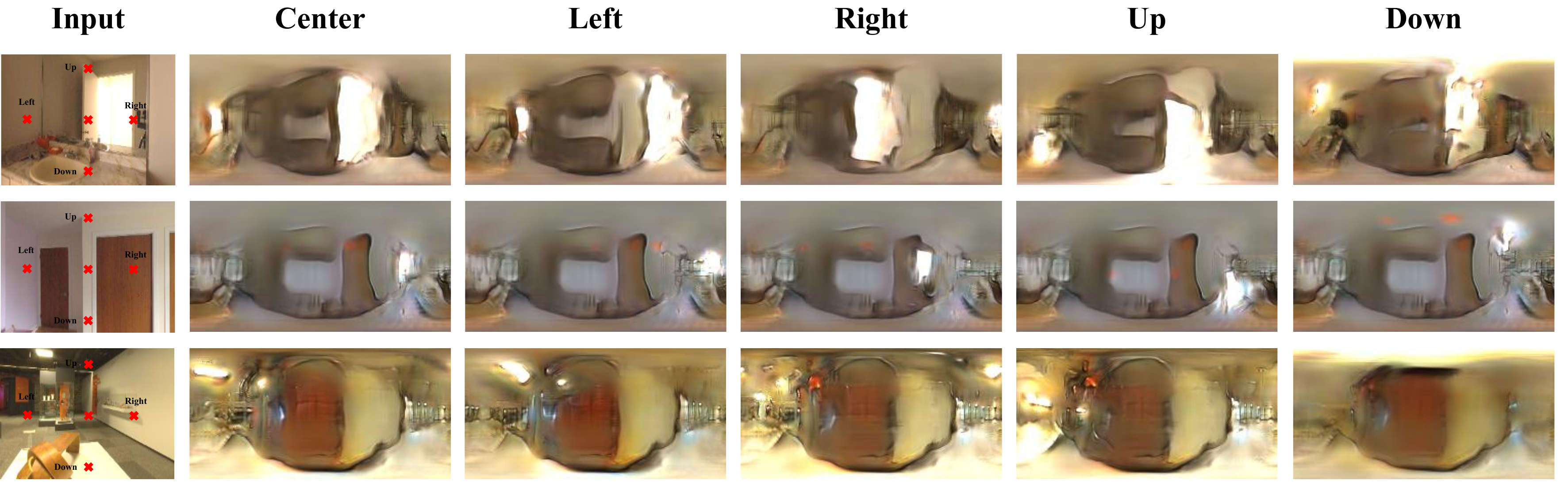}
\caption{
GMLight is capable of estimating spatially-varying illumination from different insertion positions (e.g. center, left, right, up, and down as illustrated). The samples are from the Laval Indoor HDR dataset \cite{gardner2017}.
}
\label{im_spatial}
\end{figure*}

\begin{figure}[ht]
\centering
\includegraphics[width=1.0\linewidth]{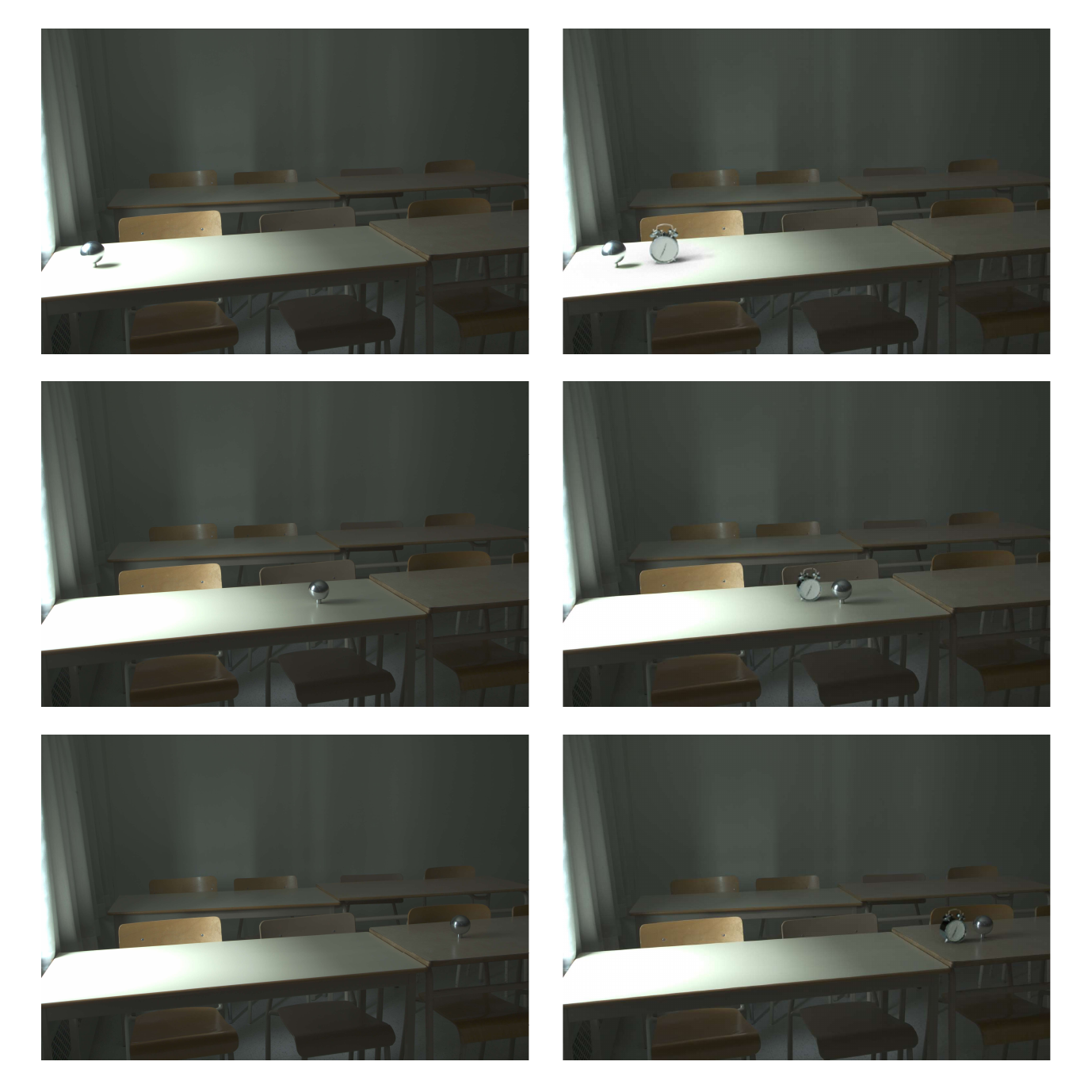}
\caption{
Illustration of spatially-varying illumination estimation by GMLight: Images in column 1 are background images \cite{garon2019fast} for 3D object insertion, where the silver sphere in each image shows real illuminations at one specific position. Images in column 2 show object relighting at different insertion positions. With GMLight-estimated illuminations, the relighting at different positions is well aligned with that of the silver spheres.
}
\label{im_spatial_render}
\end{figure}

\subsection{Quantitative Evaluation}

We conduct quantitative comparison between GMLight and other SOTA lighting estimation methods as shown in Table \ref{tab_compare}.
Specifically, each compared method predicts 200 illumination maps from the testing set to render the testing scene (three spheres of different materials).
The experimental results is tabulated in Table \ref{tab_compare}, where \textit{D}, \textit{S} and \textit{M} denote diffuse, matte silver and mirror material objects, respectively. 
As can be observed in Table \ref{tab_compare}, GMLight consistently outperforms all compared methods under different evaluation metrics and materials. EMLight \cite{zhan2021emlight} simplifies the light distribution of scenes to be spherical, ignoring the complex scene geometry. GMLight introduces depth to model scene geometry which leads to more accurate illumination estimation. Gardner \emph{et al.} \cite{gardner2017} generate illumination maps directly, but it tends to overfit training data due to the unconstrained nature of illumination estimation from a single image. Gardner \emph{et al.} \cite{gardner2019deeppara} regress the spherical Gaussian parameters of light sources, but it often loses useful frequency information and generates inaccurate shading and shadows. 
Li \emph{et al.} \cite{li2019spherical} adopt spherical Gaussian functions to reconstruct the illumination maps in the spatial domain but it often loses high-frequency illumination. 
Garon \emph{et al.} \cite{garon2019fast} recover lighting by regressing spherical harmonic coefficients, but their model struggles to regress light directions and recover high-frequency information. Although Garon \emph{et al.} \cite{garon2019fast} adopt a masked L2 loss to preserve high-frequency information, it does not fully solve the problem as illustrated in Fig. \ref{im_render}. 
NeedleLight \cite{zhan2021sparse} achieves lighting estimation in both spatial and frequency domains, while it is unable to recover illumination maps with fine details as it doesn't employ a generative projector.
In contrast, GMLight firstly regresses the accurate light parameters with a regression network, followed by a generative projector to synthesize realistic yet accurate illumination maps under guidance of the regressed parameters.

\subsection{Qualitative Evaluation}

To demonstrate that the regressed parameters provide accurate guidance for the generative projector,
we visualize the Gaussian maps, lighting maps, and intensity maps in GMLight and the corresponding ground truth in Fig. \ref{im_sample}.
As shown in Fig. \ref{im_sample}, the predicted Gaussian maps indicate the light distributions in the scenes accurately, which enables the generative projector to synthesize illumination maps with accurate light directions.
We also visualize the intensity maps of the predicted lighting maps, and the plausible intensity map compared with the ground truth demonstrate that the generative projector synthesizes accurate HDR lighting maps.

We conduct qualitative comparison between GMLight and four SOTA lighting estimation methods on object insertion task as shown in Fig. \ref{im_render}. Specifically, illumination maps are predicted from the input images and the inserted objects are rendered with the predicted illumination maps.
As illustrated in Fig. \ref{im_render}, the illumination maps predicted by GMLight present realistic and accurate light sources with fine details, thus enabling to render object with plausible shading and shadow effects.
On the other hand, Gardner \emph{et al.} \cite{gardner2017} generate the HDR illumination maps directly which make it hard to synthesize accurate light sources;
Gardner \emph{et al.} \cite{gardner2019deeppara} and Li \emph{et al.} \cite{li2019spherical} regress the representative Gaussian parameters to reconstruct the illumination maps which however causes unrealistic illumination maps with losing details especially frequency information.
Garon \emph{et al.} \cite{garon2019fast} and Zhan \emph{et al.} \cite{zhan2021sparse} regress the representation parameters to recover the scene illumination but are often constrained by the limited order of representation basis, which may incur low-frequency illumination that produces weak shading and shadows effects.

Besides the testing set, we also conduct object insertion on wild images collected from the Internet as shown in Fig. \ref{im_render2}. 
The accurate illumination maps and realistic rendering results demonstrate the exceptional generalization capability of the proposed method.

\renewcommand\arraystretch{1.2}
\begin{table}[ht]
\footnotesize
\caption{
Quantitative comparison of spatially-varying illumination estimated by GMLight and state-of-the-art methods: Evaluations were performed for three insertion positions (left, center, and right) over images from the Laval Indoor HDR dataset \cite{gardner2017}. The scores of the three spheres in Fig. \ref{im_ball} are averaged to obtain the final score.
}
\renewcommand\tabcolsep{4pt}
\centering 
\begin{tabular}{l|ccc|ccc} \hline
\multirow{2}{*}{\textbf{Methods}} & 
\multicolumn{3}{c|}{\textbf{RMSE} $\downarrow$} & 
\multicolumn{3}{c}{\textbf{si-RMSE} $\downarrow$}
\\
\cline{2-7}
 & Left & Center & Right & Left & Center & Right  \\\hline

Gardner \emph{et al.} \cite{gardner2017} & 0.168 & 0.176 & 0.171   & 0.148 & 0.159 & 0.152   \\

Gardner \emph{et al.} \cite{gardner2019deeppara} & 0.102  & 0.114 & 0.104     & 0.085 & 0.097 & 0.087   \\

Garon \emph{et al.} \cite{garon2019fast}  & 0.186 & 0.199 & 0.182    & 0.174 & 0.184  & 0.173  \\

GMLight      & \textbf{0.059} & \textbf{0.066}  &  \textbf{0.058}     & \textbf{0.043} & \textbf{0.051} & \textbf{0.041}   \\\hline
\end{tabular}
\label{tab_spatial}
\end{table}

\subsection{Spatially-varying Illumination}

Spatially-varying illumination prediction aims to recover the illumination at different positions of a scene. Fig. \ref{im_spatial} shows the spatially-varying illumination maps that are predicted at different insertion positions (center, left, right, up, and down) by GMLight. It can be seen that GMLight estimates illumination maps of different insertion positions nicely, largely due to the auxiliary depth branch that estimates scene depths and recovers the scene geometry accurately. We also evaluate spatially-varying illumination estimation quantitatively and Table \ref{tab_spatial} shows experimental results. We can see that GMLight outperforms other methods consistently in all insertion positions. The superior performance is largely attributed to the accurate geometry modeling of light distributions with scene depths.

Fig. \ref{im_spatial_render} illustrates the 3D insertion results with the estimated spatially-varying illuminations. The sample images are from \cite{garon2019fast}, where a silver sphere is employed to indicate spatially-varying illuminations at different scene positions which serve as references for evaluating the realism of 3D insertion. As Fig. \ref{im_spatial_render} shows, the inserted objects (clock) at different positions present consistent shading and shadow effect with the silver sphere, demonstrating the high-quality estimation of spatially-varying illuminations by GMLight.

\renewcommand\arraystretch{1.2}
\begin{table*}[t]
\small 
    \caption{Ablation study of the proposed GMLight. SG and GD denote spherical Gaussian representation and our proposed geometric distribution representation of illumination maps. L2 and GML denote L2 loss and geometric mover's loss that are used in the regression of light parameters. GP denotes our proposed generative projector. `GD+GML+GP' denotes the standard GMLight.
}
\renewcommand\tabcolsep{5.5pt}
\centering 
\begin{tabular}{l||ccc||ccc||ccc||ccc||c} \hline
\multirow{2}{*}{\textbf{Models}} & 
\multicolumn{3}{c||}{\textbf{RMSE} $\downarrow$} & 
\multicolumn{3}{c||}{\textbf{si-RMSE} $\downarrow$} &
\multicolumn{3}{c||}{\textbf{Angular Error} $\downarrow$} &
\multicolumn{3}{c||}{\textbf{User Study} $\uparrow$} & {\textbf{GMD} $\downarrow$}
\\
\cline{2-14}
 & D & S & M & D & S & M& D & S & M &  D & S & M  & N/A \\\hline

\textbf{SG+L2} & 0.204 & 0.213 & 0.238   & 0.188 & 0.203 & 0.229     & 9.18$^{\circ}$  & 9.42$^{\circ}$  & 9.73$^{\circ}$   & 26.0\% & 22.5\% & 18.0\%   & 5.631 \\

\textbf{GD+L2} & 0.133 & 0.161 & 0.178   & 0.117 & 0.132 & 0.161     & 7.60$^{\circ}$  & 7.88$^{\circ}$  & 8.12$^{\circ}$   & 30.5\% & 25.5\% & 22.0\%  &  5.303 \\

\textbf{GD+SML}     & 0.080 & 0.103 & 0.117   & 0.072 & 0.087 & 0.106      & 6.78$^{\circ}$  & 6.98$^{\circ}$  & 7.12$^{\circ}$   & 34.0\% & 31.5\% & 26.0\%  & 5.163 \\

\textbf{GD+GML}     & 0.073 & 0.091 & 0.102   & 0.062 & 0.069 & 0.092      & 6.61$^{\circ}$  & 6.85$^{\circ}$  & 7.04$^{\circ}$   & 35.5\% & 32.0\% & 25.5\% &  5.031 \\

\hline
\textbf{GD+GML+GP}         
& \textbf{0.051} & \textbf{0.064} & \textbf{0.078}   
& \textbf{0.037} & \textbf{0.049} & \textbf{0.074}
& \textbf{6.21$^{\circ}$}  & \textbf{6.50$^{\circ}$} & \textbf{6.77$^{\circ}$}
& \textbf{42.0\%} & \textbf{35.5\%} & \textbf{31.0\%} & \textbf{4.892}   \\\hline
\end{tabular}
\label{tab_ablation}
\end{table*}

\renewcommand\arraystretch{1.2}
\begin{table}[t]
\footnotesize
\caption{Ablation studies over anchor points, loss functions, and convolution operators: GMLight denotes the standard setting with 128 anchor points, geometric mover's loss (GML), and spherical convolution. We create five GMLight variants that use 64 and 196 anchor points, replace GML with cross-entropy loss (With CEL), replace spherical convolution with vanilla convolution (With VConv), and employ a fixed radius in the Gaussian function (With Fixed Radius), respectively (other conditions unchanged).}
\renewcommand\tabcolsep{4.5pt}
\centering 
\begin{tabular}{l|ccc|ccc} \hline
\multirow{2}{*}{\textbf{Methods}} & 
\multicolumn{3}{c|}{\textbf{RMSE} $\downarrow$} & 
\multicolumn{3}{c}{\textbf{si-RMSE} $\downarrow$}
\\
\cline{2-7}
 & D & S & M & D & S & M  \\\hline
\hline

\textbf{Anchor=64}       & 0.071  & 0.085 & 0.102     & 0.064 & 0.071  & 0.093    \\
\textbf{Anchor=196}      & 0.053  & \textbf{0.062} & 0.081    & \textbf{0.036} & 0.050  & 0.075    \\

\hline

\textbf{With CEL} & 0.062  & 0.074 & 0.094     & 0.055 & 0.059 & 0.078   \\

\textbf{With VConv}  & 0.056 & 0.069 & 0.082    & 0.044 & 0.054  & 0.083  \\

\textbf{With Fixed Radius}  & 0.063 & 0.071  & 0.085     & 0.047 & 0.056 & 0.081  \\

\hline

\textbf{GMLight}      & \textbf{0.051} & 0.064  &  \textbf{0.078} & 0.037 & \textbf{0.049} & \textbf{0.074}   \\\hline
\end{tabular}
\label{tab_ablation2}
\end{table}

\subsection{Ablation Study}

We developed several GMLight variants as listed in Table \ref{tab_ablation} to evaluate the effectiveness of the proposed designs. 
The baseline model is selected as SG+L2 which regresses spherical Gaussian parameters with L2 loss.
GD+L2, GD+SML, and GD+GML denote regress the geometric distribution of illumination with L2 loss, SML in EMLight \cite{zhan2021emlight}, and GML in GMLight, respectively.
GD+GML+GP denotes the standard GMLight with all proposed designs.
All variant models are employed to render 200 images of the testing scene (3 spheres) and the rendering results are evaluated by various metrics including the RMSE, si-RMSE, Angular Error, User Study, and the proposed GMD as shown in Table \ref{tab_ablation}.
It can be observed that GD+L2 outperforms SG+L2 clearly which demonstrates the effectiveness of geometric distributions for lighting representation. 
The superiority of the proposed GML is also verified as GD+GML produces better estimation than GD+L2 and GD+SML.
With the including of the generative projector (GP), the performance across all metrics is improved by a large margin which demonstrates that generative projector improves illumination realism and accuracy significantly.

We also ablate the effect of different number of anchor points as shown in Table \ref{tab_ablation2}.
It should be noted that the experimental results in RMSE and si-RMSE are averaged on three materials.
Compared with the standard 128 anchor points, the prediction performance with 64 anchor points drops slightly and increasing anchor points to 196 also doesn't bring clear performance gain. We conjecture that the larger number of parameters with 196 anchor points affects the regression accuracy negatively.
In addition, we also compare the geometric mover's loss and cross-entropy loss for distribution regression, compare the spherical convolution and vanilla convolution for panorama generation, and study the effect of adaptive radius in Gaussian projection.
We can see that GML outperforms cross-entropy loss (CEL) clearly as GML captures spatial information of geometric distributions effectively. In addition, spherical convolution performs better than vanilla convolution consistently in panoramic image generation and adaptive radius strategy in Gaussian projection also brings notably improvement to the performance.

\section{Conclusion}
\label{conclusion}

We present a lighting estimation framework GMLight which combines the merits of regression-based method and generation-based method. We formulate the illumination prediction as a distribution regression problem within a geometric space and design a geometric mover's loss to achieve accurate regression of illumination distribution by leveraging the real scene geometry. 
With the including of depth branch, the proposed method also enables effective estimation of spatially-varying illumination.
To synthesize panoramic illumination maps with fine details from the light distribution, a novel generative projector with progressive guidance is designed to adversarially generate illumination maps. 
Extensive experiments demonstrate that the proposed GMLight significantly outperforms previous methods in terms of relighting in object insertion.

\section{Acknowledgement}
This study is supported under the RIE2020 Industry Alignment Fund – Industry Collaboration Projects (IAF-ICP) Funding Initiative, as well as cash and in-kind contribution from the industry partner(s).

\bibliographystyle{IEEEtran}
\bibliography{cite}

\begin{IEEEbiographynophoto}{Fangneng Zhan}
received the B.E. degree in Communication Engineering and Ph.D. degree in Computer Science \& Engineering from University of Electronic Science and Technology of China and Nanyang Technological University, respectively.
His research interests include deep generative models and image synthesis \& manipulation.
He contributed to the research field by publishing more than 10 articles in prestigious conferences.
He also served as a reviewer or program committee member for top journals and conferences including TPAMI, TIP, ICLR, NeurIPS, CVPR, ICCV, and ECCV.
\end{IEEEbiographynophoto}

\vspace{-0.25cm}
\begin{IEEEbiographynophoto}{Yingchen Yu}
obtained the B.E. degree in Electrical \& Electronic Engineering at Nanyang Technological University, and M.S. degree in Computer Science at National University of Singapore. He is currently pursuing the Ph.D. degree at School of Computer Science and Engineering, Nanyang Technological University under Alibaba Talent Programme. His research interests include computer vision and machine learning, specifically for image synthesis and manipulation.
\end{IEEEbiographynophoto}

\vspace{-0.25cm}
\begin{IEEEbiographynophoto}{Changgong Zhang} is currently an algorithm engineer of Alibaba DAMO Academy. He received the PhD degree in Computer Graphics and Visualization from Delft University of Technology in 2017. His research interests include 3D vision, inverse rendering, and scientific visualization.
\end{IEEEbiographynophoto}

\vspace{-0.25cm}
\begin{IEEEbiographynophoto}{Rongliang Wu}
received the B.E. degree in Information Engineering from South China University of Technology, and M.S. degree in Electrical and Computer Engineering from National University of Singapore. He is currently pursuing the Ph.D. degree at School of Computer Science and Engineering, Nanyang Technological University. His research interests include computer vision and deep learning, specifically for facial expression analysis and generation.
\end{IEEEbiographynophoto}

\vspace{-0.25cm}
\begin{IEEEbiographynophoto}{Wenbo Hu}
is currently a Ph.D. candidate in the Department of Computer Science and Engineering, The Chinese University of Hong Kong. He received his B.Sc. degree in computer science and technology from Dalian University of Technology, China, in 2018. His research interests include computer vision, computer graphics and deep learning.
\end{IEEEbiographynophoto}

\begin{IEEEbiographynophoto}{Shijian Lu} is an Assistant Professor in the School of Computer Science and Engineering, Nanyang Technological University. He received his PhD in Electrical and Computer Engineering from the National University of Singapore. His research interests include computer vision and deep learning. He has published more than 100 internationally refereed journal and conference papers. Dr Lu is currently an Associate Editor for the journals of Pattern Recognition and Neurocomputing.
\end{IEEEbiographynophoto}

\begin{IEEEbiographynophoto}{Feiying Ma}
is a senior algorithm engineer at Alibaba Damo Academy and currently is in charge of openAI platform on Alibaba Cloud. Her research interests include image processing, virtual reality, augment reality, real-time rendering, and 3D reconstruction.
\end{IEEEbiographynophoto}

\begin{IEEEbiographynophoto}{Xuansong Xie} is a senior staff engineer and technical director of Damo Academy, joined Alibaba, in 2012, and currently is in charge of the Alibaba Design Intelligence and Health Intelligence team, focusing on vision generation and enhancement, image retrieval, medical image \& language intelligence, and other AI technology R\&D.
\end{IEEEbiographynophoto}

\begin{IEEEbiographynophoto}{Ling Shao} is the CEO and the Chief Scientist of the Inception Institute of Artificial Intelligence (IIAI), Abu Dhabi, United Arab Emirates. He was also the initiator and the Founding Provost and Executive Vice President of the Mohamed bin Zayed University of Artificial Intelligence (the world's first AI University), UAE. His research interests include computer vision, deep learning, medical imaging and vision and language. He is a fellow of the IEEE, the IAPR, the IET, and the BCS.
\end{IEEEbiographynophoto}

\end{document}